\documentclass[journal]{IEEEtran}
\usepackage{color}
\usepackage{multirow}
\usepackage{colortbl}
\usepackage{algorithmic}
\usepackage{algorithm}
\usepackage{amssymb}
\usepackage{blindtext}
\usepackage{marginnote}
\usepackage{makeidx}         
\usepackage{graphicx}        
\usepackage{multicol}        
\usepackage[bottom]{footmisc}
\usepackage{ifpdf}
\usepackage{cite}
\usepackage[cmex10]{amsmath}
\usepackage{algorithmic}
\usepackage{array}
\usepackage{mdwmath}
\usepackage{mdwtab}
\usepackage{eqparbox}
\usepackage{fixltx2e}
\usepackage{stfloats}
\usepackage{hyperref}
\usepackage{url}
\hypersetup{
colorlinks = true,
urlcolor = blue,
linkcolor = blue,
citecolor = blue
}
\usepackage{lipsum}
\newcommand*\rot[1][25]{\rotatebox{25}}
\IEEEoverridecommandlockouts
\begin{document}
%
\title{Bounded Fuzzy Possibilistic Method}

\author{Hossein~Yazdani
\IEEEcompsocitemizethanks{\IEEEcompsocthanksitem H. Yazdani was with the Faculty
of Electronic, the Faculty of Computer Science and Management, and the Department of Information Systems at Wroclaw University of Science and Technology, Poland, and Climax Data Pattern, Huston, Texas, USA.\protect\\
E-mail: hossein.yazdani@pwr.edu.pl, yazdanihossein@yahoo.com
}
}
\markboth{2019}%
{Shell \MakeLowercase{\textit{et al.}}: Bounded Fuzzy Possibilistic Method}
\maketitle

\begin{abstract}
This paper introduces Bounded Fuzzy Possibilistic Method (BFPM) by addressing several issues that previous clustering/classification methods have not considered. In fuzzy clustering, object's membership values should sum to 1. Hence, any object may obtain full membership in at most one cluster. Possibilistic clustering methods remove this restriction. However, BFPM differs from previous fuzzy and possibilistic clustering approaches by allowing the membership function to take larger values with respect to all clusters. Furthermore, in BFPM, a data object can have full membership in multiple clusters or even in all clusters. BFPM relaxes the boundary conditions (restrictions) in membership assignment. The proposed methodology satisfies the necessity of obtaining full memberships and overcomes the issues with conventional methods on dealing with overlapping. Analysing the objects' movements from their own cluster to another (mutation) is also proposed in this paper. BFPM has been applied in different domains in geometry, set theory, anomaly detection, risk management, diagnosis diseases, and other disciplines. Validity and comparison indexes have been also used to evaluate the accuracy of BFPM. BFPM has been evaluated in terms of accuracy, fuzzification constant (different norms), object's movement analysis, and covering diversity. The promising results prove the importance of considering the proposed methodology in learning methods to track the behaviour of data objects, in addition to obtain accurate results.   \\[0.5 cm]
\textit{Index Terms$\bf -$}Bounded Fuzzy Possibilistic Method, Membership function, Critical object, Object movement, Mutation, Overlapping, Similarity function, Weighted Feature Distance, Supervised learning, Unsupervised learning,  Clustering. \\[0.3cm]
\end{abstract}

\IEEEpeerreviewmaketitle
\section{{\textbf{Introduction}}}
\IEEEPARstart{C}{lustering} is a form of unsupervised learning that splits data into different groups or clusters by calculating the similarity between objects contained in a dataset \cite{one}. 
More formally, assume that we have a set of $n$ objects represented by $ O = \{ o_1, o_2, \;  ... \; , o_n \}$ in which each object is typically described by numerical $feature-vector$ data that has the form $X \;= \; \{ x_1,... \;, x_d\} \; \subset R^d $, where $d$ is the dimension of the search space or the number of features \cite{two}. A cluster is a set of objects, and the relation between clusters and objects is often represented by a matrix with values $\{ u_{ij}\}$, where $u$ represents a membership value, $j$ is the $j^{th}$ object in the dataset, and $i$ is the $i^{th}$ cluster \cite{three}. A partition or membership matrix is often represented as a  $c\times n$ matrix $U = [u_{ij}]$, where $c$ is the number of clusters \cite{four}. Crisp, fuzzy and possibilistic are three types of partitioning methods \cite{five}. Crisp clusters are non-empty, mutually-disjoint subsets of $O$:
\begin{equation}
\nonumber
\label{Crisp partition}
M_{hcn} = \bigg\lbrace U \; \in \Re^{c\times n}| \; u_{ij} \; \in \lbrace 0,1 \rbrace, \; \;  \forall \; i,j;
\end{equation}
\begin{equation}
 0 < \sum_{j=1}^n u_{ij} < n,  \; \;  \forall i; \; \;  \sum_{i=1}^c u_{ij} =1, \; \;  \forall j  \; \; \bigg\rbrace
\end{equation}\\
where $u_{ij}$ is the membership of object $o_j$ in cluster $i$. If the object $o_j$ is a member of cluster $i$, then $u_{ij} \;  = \; 1;$ otherwise, $u_{ij} \;  = \; 0$. 
Fuzzy clustering is similar to crisp clustering \cite{six}, but each object can have partial memberships in more than one cluster \cite{seven} (to cover overlapping). This condition is stated by Eq. (\ref{fuzzy partition}), where an object may obtain partial nonzero memberships in several clusters, but only a full membership in one cluster.
\begin{equation}
\nonumber
\label{fuzzy partition}
M_{fcn} = \bigg\lbrace U \; \in \Re^{c\times n}| \; u_{ij} \; \in [ 0,1 ], \; \; \forall \; i,j;
\end{equation}
\begin{equation}
0 < \sum_{j=1}^n u_{ij} < n, \;\; \forall i; \;\;  \sum_{i=1}^c u_{ij} =1, \;\; \forall j \; \; \bigg\rbrace
\end{equation}\\
According to Eq. (\ref{fuzzy partition}), each column of the partition matrix must sum to 1 $(\sum_{i=1}^c u_{ij} =1)$. Thus, a property of fuzzy clustering is that, as $c$ becomes larger, the $u_{ij}$ values must become smaller \cite{eight}.
An alternative partitioning approach is \textit{possibilistic clustering} \cite{nine}. In Eq. (\ref{posibilistic-partitioning}), the condition $( \sum_{i=1}^c u_{ij} =1 )$ is relaxed by substituting it with $( \underset{1 \leq i \leq c}{max} \; \; u_{ij} >0)$.
\begin{equation}
\nonumber
\label{posibilistic-partitioning}
M_{pcn} = \bigg\lbrace U \; \in \Re^{c\times n}| \; u_{ij} \;\; \in [ 0,1 ], \;\;  \forall \; i,j;
\end{equation}
\begin{equation}
0 < \sum_{j=1}^n u_{ij} \leq n, \; \forall i; \;\;  \max_{1\leq i\leq c}\; u_{ij} >0, \;\; \forall j \; \; \bigg\rbrace
\end{equation}
Based on Eq. (\ref{Crisp partition}), Eq. (\ref{fuzzy partition}), and Eq. (\ref{posibilistic-partitioning}), it is easy to see that all crisp partitions are subsets of fuzzy partitions, and a fuzzy partition is a subset of a possibilistic partition, i.e., $M_{hcn}\subset M_{fcn}\subset M_{pcn}$. Possibilistic method has some drawbacks such as offering trivial null solutions \cite{ten}, and the method needs to be tuned in advance as the method strongly depends on good initialization steps \cite{eleven}. AM-PCM (Automatic Merging Possibilistic Clustering Method) \cite{twelve}, Graded possibilistic clustering \cite{thirteen}, and some other approaches such as soft transition techniques \cite{fourteen} are proposed to cover the issues with possibilistic methods. 
\subsection{\bf FCM Algorithm}
In prototype-based (centroid-based) clustering, the data is described by a set of \textit{point prototypes} in the data space \cite{fifteen}. There are two important types of prototype-based FCM algorithms. One type is based on the fuzzy partition of a sample set \cite{sixteen} and the other is based on the geometric structure of a sample set in a kernel-based method \cite{seventeen}. 
The FCM function may be defined as \cite{one}:
\begin{equation}
\label{General-FCM}
J_m(U,V) \; = \; \sum_{i=1}^c \sum_{j=1}^n \; u_{ij}^m \;|| X_j - V_i||_{A} ^2 \; ;
\end{equation}
where U is the $(c \times n)$ partition matrix, $ V= \{ v_1, v_2,..., v_c \} $ is the vector of $c$ cluster centers (prototypes) in $ \Re^d , \; m  > 1 $ is the fuzzification constant, and $ ||.||_A $ is any inner product A-induced norm \cite{eighteen}, i.e., $ ||X||_{A} = \sqrt{X^T AX}$ or the distance function such as Minkowski distance, presented by Eq. (\ref{distance equation}). Eq.(\ref{General-FCM}) \cite{nineteen} makes use of Euclidean distance function by assigning $(k =2)$ in Eq. (\ref{distance equation}). \\
\begin{equation}
\label{distance equation}
d_k(x,y)= \bigg(\sum_{j=1}^d \mid x_j - y_j\mid ^k \bigg)^{(1/k)}
\end{equation}
In this paper, the Euclidean norm $(A=I)$ or $(k=2)$ is used for the experimental verifications, although there are some issues with conventional similarity functions to cover diversity in their feature spaces \cite{twenty}.
\subsection{\bf Kernel FCM}
In kernel FCM, the dot product $ \Big{(} \phi(X).\phi(X) = k(X,X) \Big{)}$ is used to transform feature vector $X$, for non-linear mapping function $\Big{(}\phi : \; X \longrightarrow \phi(X) \; \; \in  \; \; \Re^{D_k} \Big{)}$, where $D_k$ is the dimensionality of the feature space. Eq. (\ref{kFCM}) presents a non-linear mapping function for Gaussian kernel \cite{twenty one}.\\
\begin{equation}
\label{kFCM}
J_m(U;k) = \sum_{i=1}^c \Big{(} \sum_{j=1}^n \sum_{k=1}^n (u_{ij}^m u_{ik}^m \; \; d_k(x_j,x_k))/2 \sum_{l=1}^n u_{il}^m \Big{)} 
\end{equation}\\
where $U \in M_{fcn}, \;\; m > 1$ is the fuzzification parameter, and  $d_k(x_j,x_k)$ is the kernel base distance \cite{twenty two}, replaces Euclidean function, between the $j^{th}$ and $k^{th}$ feature vectors:\\
\begin{equation}
\label{kernel distance}
d_k(x_j,x_k) = k(x_j,x_j) + k(x_k,x_k) - 2k(x_j,x_k)
\end{equation}\\
The other sections in this paper are organized as follows. In Section \ref{Issues}, challenges on conventional clustering and membership functions are discussed. To clarify the challenges, numerical examples are explored. Section \ref{BFPCM Method} introduces BFPM methodology on membership assignments and discusses how the method overcomes the issues with conventional methods. This section also presents an algorithm for clustering problems using BFPM methodology. Experimental results and cluster validity functions are discussed in Section \ref{Exp_result}. Discussions and conclusions are presented in Section \ref{Conclusion}.
\section{\textbf{Challenges on Conventional Methods}}
\label{Issues}
Uncertainty is the main concept in fuzzy type-I, fuzzy type-II, probability, possibilistic, and other methods \cite{twenty three}. To get a better understanding of the issues on conventional membership functions that deal with uncertainty, several examples in wider perspectives are presented in the following sections. The necessity of considering a comprehensive membership function on data objects is being brighten when intelligent systems are used to implement mathematical equations in uncertain conditions. The aim of this paper is to remove restrictions on data objects to participate in as much clusters as they can \cite{twenty four}. Lack of this consideration in learning methods weakens the accuracy and the capability of the proposed methods \cite{twenty five}. The following examples explore the issues on conventional methods in two dimensional search space. In higher dimensional search spaces, Big data, and social networks where overlapping is the key role \cite{twenty six}, \cite{twenty seven}, the influences of miss-assignments massively skew the final results. Examples are selected from geometry, set theory, medicine, and other domains to highlight the importance of considering objects' participation in more clusters.
\subsection{\bf Example from Geometry:}
Assume $ U_G = \Big{\lbrace} u_{ij}(x) \Big{|} \; x_j \in L_i  \; ,  \; u_{ij} \rightarrow [0,1] \Big{\rbrace}$ is a membership function that assigns a membership degree to each point $x_j$ with respect to each line $L_i$, where each line represents a cluster. Some believe that a line cannot be a cluster, but we should note that data pattern or data model can be represented as a function in different norms, which in here, the data models can be presented as lines in two dimensional search space. Now consider the following equation which describes $c$ lines crossing at the origin:
\begin{equation}
AX =0
\label{transversal-equ}
\end{equation}
where matrix $A$ is a $c \times d$ coefficient matrix, and $X$ is a $d \times 1$ matrix, in which $c$ is the number of lines and $d$ is the number of dimensions (in this example is 2). From a geometrical point of view, each line containing the origin is a subspace of $R^d$. Eq. (\ref{transversal-equ}) describes $c$ lines as subspaces. Without the origin, each of those lines is not a subspace, since the definition of a subspace comprises the existence of the null vector as a condition in addition to other properties \cite{twenty eight}. When trying to design a fuzzy-based \cite{twenty nine} clustering method that can create clusters using the points in all lines, it should be noted that removing or decreasing a membership value, with respect to the origin of each cluster, ruins the subspace.
\begin{equation}
\label{transversal-equ}
\begin{bmatrix}
C_{1,1} & C_{1,2}\\
C_{2,1} & C_{2,2}\\
\vdots & \vdots  \\
C_{c1} & C_{c,2}\\
\end{bmatrix}
\times
\begin{bmatrix}
X_1\\
X_2\\
\end{bmatrix}
=
\begin{bmatrix}
0\\
0\\
\vdots \\
0\\
\end{bmatrix}
\end{equation}
For instance, $x_1=0 $ , $x_2=0$ , $x_1=x_2$ , and $x_1 = -x_2$ are equations representing some of those lines shown by Eq. (\ref{lines-equ}) with infinite data objects (points) on them. 
\begin{equation}
\label{lines-equ}
\begin{bmatrix}
1 & 0\\
0 & 1\\
1 & 1  \\
1 & -1\\
\end{bmatrix}
\times
\begin{bmatrix}
X_1\\
X_2\\
\end{bmatrix}
=
\begin{bmatrix}
0\\
0\\
0 \\
0\\
\end{bmatrix}
\end{equation}
To clarify the idea, just two of those lines $A: \{ x_2=0 \}$ and $B: \{x_1=0\}$ with five points on each, including the origin, are selected for this example.
\begin{equation}
\label{Lines}
\nonumber
	A = \lbrace p_{11}, p_{12}, p_{13}, p_{14}, p_{15} \rbrace  
\end{equation}
\begin{equation}
\label{Lines}
\nonumber
	= \lbrace (-2,0), (-1,0), (0,0), (1,0), (2,0) \rbrace 
\end{equation}
\begin{equation}
\label{Lines}
\nonumber
	U_A = \Big{\lbrace} u_{A}(p_j) \Big{|} p_j \in L_1 \subset \Re^2 \Big{\rbrace}  
\end{equation}\\
%
\begin{equation}
\nonumber
 B = \lbrace p_{21}, p_{22},p_{23}, p_{24}, p_{25}\rbrace 
\end{equation}
%
\begin{equation}
\nonumber
= \lbrace (0,-2), (0,-1), (0,0), (0,1), (0,2) \rbrace 
\end{equation}
\begin{equation}
\label{Lines}
\nonumber
	U_B = \Big{\lbrace} u_{B}(p_j) \Big{|} p_j \in L_2 \subset \Re^2 \Big{\rbrace} \\ 
\end{equation}\\
where $p_{j}=(x_1,x_2)$. 
The origin is a member of all lines, but for convenience, it has been given different names such as $p_{13}$ and $p_{23}$ in each line above. 
The point distances with respect to each line and Euclidean function $ \bigg{(} d_k(x,y)= \big(\sum_{j=1}^d \mid x_j - y_j\mid ^2 \big)^{(1/2)} \bigg{)} $ are shown in the $(2 \times 5)$ matrices below, where $2$ is the number of clusters and $5$ is the number of objects.
\begin{minipage}{0.5\textwidth}
 \begin{flushleft}
 \begin{center}
 \begin{equation}
\label{similarity Matrix}
\nonumber
D_1 =
\begin{bmatrix}
0.0 & , & 0.0 & , & 0.0 & , & 0.0 & , & 0.0 \\
2.0 & , & 1.0 & , & 0.0 & , & 1.0 & , & 2.0\\
\end{bmatrix}
\end{equation}
\end{center}
 \end{flushleft}
\end{minipage}
\begin{minipage}{0.5\textwidth}
\begin{flushright}
\begin{center}
\begin{equation}
\label{similarity Matrix}
\nonumber
D_2=
\begin{bmatrix}
2.0 & , & 1.0  & , & 0.0 & , & 1.0 & , & 2.0 \\
0.0 & , & 0.0  & , & 0.0 & , & 0.0 & , & 0.0\\
\end{bmatrix}
\end{equation}
\end{center}
\end{flushright}    
\end{minipage} \\[0.5cm]
A zero value in the first matrix in the first row indicates that the object is on the first line. For example in $D_1$, the first row shows that all the members of set $A$ are on the first line, and the second row shows how far each one of the points on the first line are from the second line (cluster). Likewise the matrix $D_2$ shows the data points on the second line.  
\begin{figure}[!h]
\begin{center}
\includegraphics[width=6cm,height=3.5cm]{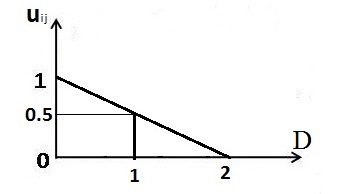}
\caption{A sample membership function for a fuzzy input variable "distance" for points (objects) with respect to each line (cluster).}
\label{figure1}
\end{center}
\end{figure}
Membership values are assigned to each point, using crisp and fuzzy sets as shown in the matrices below by using an example of membership function presented by Eq. {(\ref{Mem_Matrix}) with respect to Fig. \ref{figure1} for crisp and fuzzy methods, besides considering the conditions for these methods} described by Eq. (\ref{Crisp partition}) and Eq. (\ref{fuzzy partition}). 
\begin{equation}
\label{Mem_Matrix}
u_{ij} = \left\{
\begin{array}{l l}
1 & \quad \mbox{if $d_{ij} = 0$ }\\
\mbox{$ 1- \frac{d_{ij}}{d_{\delta}} $ } & \quad \mbox{ if $ 0 < d_{ij} \leqslant d_{\delta}$ }\\
0 & \quad \mbox{if $d_{ij} > d_{\delta}$}\\
\end{array} \right.
\end{equation}
where $d_{ij}$ is the Euclidean distance of object $x_{j}$ from cluster $i$, and $d_{\delta}$ is a constant that is used to normalize the values. In this example $(d_{\delta}=2)$. \\
\begin{picture}(10.3,5.0)
\linethickness{0.3mm}
\put(0,0){\line(1,0){245}}
\end{picture}\\
\begin{minipage}{0.5\textwidth}
 \begin{flushleft}
 \begin{center}
\begin{equation}
\label{Mem_Matrix_L1}
\nonumber
U_{crisp}(A) =
\begin{bmatrix}
1.0 &,& 1.0 &, & {\bf 1.0} & , & 1.0 & , & 1.0 \\
0.0 &, & 0.0 & , & {\bf 0.0} & , & 0.0 & ,& 0.0\\
\end{bmatrix}
\end{equation}
\end{center}
 \end{flushleft}
\end{minipage}
\begin{minipage}{0.5\textwidth}
 \begin{flushright}
 \begin{center}
\begin{equation}
\label{Mem_Matrix_L2}
\nonumber
U_{crisp}(B) =
\begin{bmatrix}
0.0 & , & 0.0 & , & {\bf 0.0} & , & 0.0 & , & 0.0 \\
1.0 & , & 1.0 & , & {\bf 0.0} & , & 1.0 & , & 1.0\\
\end{bmatrix}
\end{equation}
\end{center}
 \end{flushright}
\end{minipage}
or\\
\begin{minipage}{0.5\textwidth}
 \begin{flushleft}
 \begin{center}
\begin{equation}
\label{Mem_Matrix_L1}
\nonumber
U_{crisp}(A) =
\begin{bmatrix}
1.0 &,& 1.0 &, & {\bf 0.0} & , & 1.0 & , & 1.0 \\
0.0 &, & 0.0 & , & {\bf 0.0} & , & 0.0 & ,& 0.0\\
\end{bmatrix}
\end{equation}
\end{center}
 \end{flushleft}
\end{minipage}
\begin{minipage}{0.5\textwidth}
 \begin{flushright}
 \begin{center}
\begin{equation}
\label{Mem_Matrix_L2}
\nonumber
U_{crisp}(B) =
\begin{bmatrix}
0.0 & , & 0.0 & , & {\bf 0.0} & , & 0.0 & , & 0.0 \\
1.0 & , & 1.0 & , & {\bf 1.0} & , & 1.0 & , & 1.0\\
\end{bmatrix}
\end{equation}
\end{center}
 \end{flushright}
\end{minipage}
\begin{center}
  - - - - - - - - - - - - - - - - - - - - - - - - - - - - - - - - - - -
\end{center}
%
%
%
\begin{minipage}{0.5\textwidth}
 \begin{flushleft}
 \begin{center}
\begin{equation}
\label{Mem_Matrix_L1}
\nonumber
U_{fuzzy}(A) =
\begin{bmatrix}
1.0 & , & 0.5 & , & {\bf 0.5} & , & 0.5 & , & 1.0 \\
0.0 & , & 0.5 & , & {\bf 0.5} & , & 0.5 & , & 0.0\\
\end{bmatrix}
\end{equation}
\end{center}
 \end{flushleft}
\end{minipage}
\begin{minipage}{0.5\textwidth}
 \begin{flushright}
 \begin{center}
\begin{equation}
\label{Mem_Matrix_L2}
\nonumber
U_{fuzzy}(B) =
\begin{bmatrix}
0.0 & , & 0.5 & , & {\bf 0.5} & , & 0.5 & , & 0.0 \\
1.0 & , & 0.5 & , & {\bf 0.5} & , & 0.5 & , & 1.0\\
\end{bmatrix}
\end{equation}
\end{center}
 \end{flushright}
\end{minipage}
\begin{picture}(10.3,5.0)
\linethickness{0.3mm}
\put(0,0){\line(1,0){245}}
\end{picture}\\[0.2cm]
By selecting crisp membership functions in membership assignments, the origin can be a member of just one cluster and must be removed from other clusters. Given the properties of fuzzy membership functions, if the number of clusters increases, the membership value assigned to each object will decrease proportionally. For instance, in the case of two clusters, the membership of the origin is $u_{13} =1/2$, but if the number of clusters increases to $c$ using Eq. (\ref{transversal-equ}) we obtain $u_{13} = \frac{1}{c}$.
Hence, a point $p_{ij}$ will have a membership value $(u_{ij} ) $ that is smaller than the value that we can expect to get intuitively $E(u_{ij})$ {(typicality \cite{nine} value for $(u_{ij})$)}. For instance $( E(u_{13}) = E(u_{12}) = 1)$ given that they are points in the line. However, as the number of clusters increases, we will obtain smaller values for $(u_{ij} )$ since ($ 1 < c \;\; \longmapsto \; \; \; u_{ij} < E(u_{ij}) $ ; $1 \ll c  \; \; \longmapsto  \;\; \; u_{ij} \mapsto 0$). Possibilistic approaches allow data objects to obtain larger values in membership assignments. But PCM (Possibilistic Clustering Method) needs a good initialization to provide accurate clustering \cite{ten}. According to PCM condition $(u_{ij} \geq 0)$ the trivial null solutions should be handled by modifications on membership assignments.
\subsection{\bf Examples from Set Theory}
\begin{figure}[!h]
\begin{center}
\includegraphics[width=6.8cm,height=4.2cm]{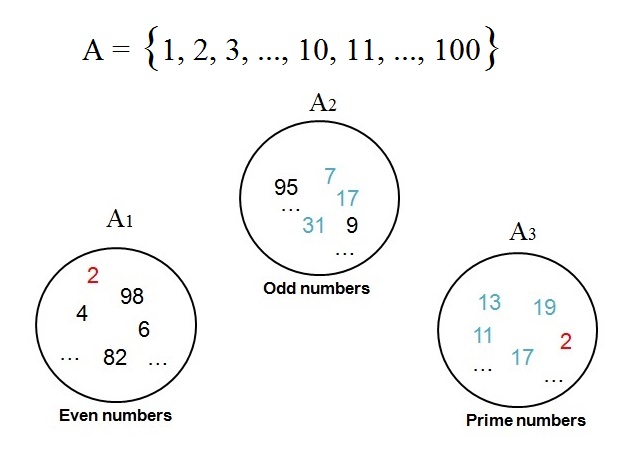}
\caption{Categorizing a set of natural numbers into some categories (even ($A_1$), odd ($A_2$), or prime numbers ($A_3$)), to evaluate the arithmetic operations (the intersection $A_1 \cap A_2$ and the union $A_1 \cup A_2$) by learning methods.}
\label{settheory}
\end{center}
\end{figure}

\begin{figure}[!h]
\begin{center}
\includegraphics[width=6.2cm,height=4.0cm]{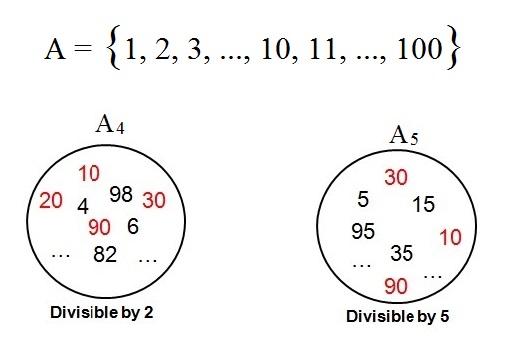}
\caption{Categorizing a set of natural numbers into some categories (divisible by two ($A_4$) or five ($A_5$)), to evaluate the arithmetic operations (the intersection and the union) by learning methods.}
\label{divisible}
\end{center}
\end{figure}

In set theory, we can extract some subsets from a superset according to the properties of members. Usually, members can participate in more than one set (cluster). For instance, $ ( A= \Big{ \lbrace } x  \Big{|} \; x \in N \; \& \; x \leq 100 \Big{ \rbrace } )$ a set of natural numbers $(\bf{N} \subset \Re)$ can be categorized into different subsets: "Even" $(A_1 = \Big{ \lbrace } x  \Big{ | } \; x \in A \; \& \; x \; mod \;  2=0 \Big{ \rbrace })$, "Odd" $( A_2 = \Big{ \lbrace } x  \Big{ | } \; x \in A \; \& \; x \; mod \;  2=1 \Big{ \rbrace } )$, and "Prime" $ ( A_3 = \Big{ \lbrace } x  \Big{ | }  \; x \in A  : \; \forall a,b \in N, x| ab \Rightarrow  ( x=a \vee x=b) \Big{ \rbrace } )$ numbers, presented by Fig. \ref{settheory}. According to set theory, we can distinguish some members that participate in other subsets with full memberships. The other example is to categorize the set $(A)$ into two clusters: numbers divisible by two $( A_4 = \Big{ \lbrace } x \Big{|} \; x \in A \; \& \; x \; mod \;  2=0 \Big{ \rbrace } ) $ and $ ( A_5 = \Big{ \lbrace } x \Big{|} \; x \in A \; \& \; x \; mod \;  5=0 \Big{ \rbrace } ) $ numbers divisible by five, presented by Fig. \ref{divisible}. By considering the examples, we see some members can participate in more clusters, and we cannot remove any of them form any of those sets. In other words, we cannot restrict objects (members) to participate in only one cluster, or partially participate in other clusters. 

Removing or restricting objects from their participation in other clusters leads to losing very important information, and consequently weakens the accuracy of learning methods. Members should be treated by a comprehensive method that allows objects to participate in more, even all clusters as full members with no restriction.
From Fig. \ref{divisible}, this is obvious that we cannot say the member $30$ is only a member of cluster "divisible by 2", or is a half member of this cluster. Member $30$ is a full member of both clusters, which cannot be precisely explored by conventional methods. More formally, for a set of objects that should be clustered into two clusters $A^{'}$ and $A^{''}$, crisp methods can cover the union $(A^{'} \cup A^{''})$, means all objects are categorized in clusters, but cannot provide the intersection $(A^{'} \cap A^{''})$, means that no object can participate in more than one cluster even for the mandatory cases such as presented examples from geometry and set theory. Other conventional methods reduce the memberships assigned to objects that participate in more than one cluster, which means objects cannot be full members of different clusters.
\subsection{\bf Examples from other Domains}
Another example is a  student from two or more departments that needs to be assigned full memberships from more than one department, as a top student based on their participations. Assume we plan to categorize a number of students (S) into some clusters, based on their skills with respect to each department or course \textit{D} (mathematics, physics, chemistry and so on) which can be presented as $U_D = \Big{\lbrace }  u_{ij}(s) \Big{|} s_j \in D_i \; , \; u_{ij} \rightarrow [0,1] \Big{ \rbrace }$. Again, assume that a very good student (\textit{$s_j$}) with potential ability to participate in more than one cluster ($D_i$) with the full membership degree (1), as follow:
\begin{center}
 	$u_{1j}  \; = \; 1$   ,   $u_{2j} \; = \; 1$\\ , ... ,   and  $u_{nj} \; = \; 1$
\end{center}
As the equations show, $s_j$ is a full member of some clusters (departments) $D_1, D_2, ..., D_n$. The following membership assignments are calculated based on crisp and fuzzy methods for membership values for student $s_1$ who is a member of two departments $D_1$ and $D_2$ out of four:  \\
\begin{equation}
\label{Mem_Matrix_T1}
\nonumber
u_{ij} =
\begin{bmatrix}
u_{1j} & , & u_{2j} & , & u_{3j} & , & u_{4j} &   \\
\end{bmatrix}
\end{equation}
\begin{picture}(10.3,5.0)
\linethickness{0.3mm}
\put(0,0){\line(1,0){245}}
\end{picture}\\
\begin{minipage}{0.5\textwidth}
 \begin{flushleft}
 \begin{center}
\begin{equation}
\label{Mem_Matrix_T1}
\nonumber
U_{Crisp}(s_1) =
\begin{bmatrix}
{\bf 1.0} & , & 0.0 & , & 0.0 & , & 0.0 &   \\

\end{bmatrix}
\end{equation}
\end{center}
 \end{flushleft}
\end{minipage}
 or \\
\begin{minipage}{0.5\textwidth}
 \begin{flushright}
 \begin{center}
\begin{equation}
\label{Mem_Matrix_T2}
\nonumber
U_{Crisp}(s_1) =
\begin{bmatrix}
0.0 & , & { \bf 1.0} & , & 0.0 & , & 0.0 & \\
\end{bmatrix}
\end{equation}
\end{center}
 \end{flushright}
\end{minipage}\\
\begin{center}
 - - - - - - - - - - - - - - - - - - - - - - - - - - - - - - - - -
\end{center}
\begin{equation}
\label{Mem_Matrix_L1}
\nonumber
U_{fuzzy}(s_1) =
\begin{bmatrix}
{\bf 0.5} & , & {\bf 0.5} & , & 0.0 & , & 0.0 &  \\
\end{bmatrix}
\end{equation}
\begin{picture}(10.3,5.0)
\linethickness{0.3mm}
\put(0,0){\line(1,0){245}}
\end{picture}\\
As results show, students can obtain credit from just one department in crisp method. In fuzzy method, their memberships and credits are divided into the number of departments. In all domains and disciplines, we need to remove the limitations and restrictions in membership assignments to allow objects to participate in more clusters to track their behaviour. In medicine, we need to evaluate individuals before being affected to any disease categories to cut the further costs for treatments \cite{thirty}. In such a case, we need to evaluate individuals on their participations in other clusters without any restriction to track their potential ability to participate in other clusters. 
\section{\textbf{New Learning Methodology (BFPM)}}
\label{BFPCM Method}
Bounded Fuzzy Possibilistic Method (BFPM) makes it possible for data objects to have full memberships in several or even in all clusters, in addition to provide the properties of crisp, fuzzy, and possibilistic methods. This method also overcomes the issues on conventional clustering methods, in addition to facilitate objects' movements (mutation) analysis. This is indicated in Eq. (\ref{flexible fuzzy partition}) with the normalizing condition of $1/c \sum_{i=1}^c u_{ij}$. BFPM allows objects to participate in more even all clusters in order to evaluate the objects' movement from their own cluster to other clusters in the near future. Knowing the exact type of objects and studying the object's movement in advance is very important for crucial systems. \\
%
\begin{equation}
\nonumber
\label{flexible fuzzy partition}
M_{bfpm} = \bigg\lbrace U \; \in \Re^{c\times n}| \; u_{ij} \; \in [0,1], \; \; \forall \; i,j;
\end{equation}
\begin{equation}
 0 < \sum_{j=1}^n u_{ij} \leq n, \;\; \forall i; \;\; 0< \; 1/c \sum_{i=1}^c u_{ij} \leq 1, \;\; \forall j  \; \; \bigg\rbrace
\end{equation}\\
%
BFPM avoids the problem of reducing the objects' memberships when the number of clusters increases. According to the geometry example, objects (points) are allowed to obtain full memberships from more than one even all clusters (lines) with no restrictions \cite{thirty one}. Based on the membership assignments provided by BFPM we can obtain the following results for the points on the lines.\\

%
\begin{equation}
\label{Mem_Matrix_L1}
\nonumber
U_{bfpm}(A) =
\begin{bmatrix}
1.0 & , & 1.0 & , & {\bf 1.0} & , & 1.0 & , & 1.0 \\
0.0 & , & 0.5 & , & {\bf 1.0} & , & 0.5 & , & 0.0\\
\end{bmatrix}
\end{equation}\\
%
\begin{equation}
\label{Mem_Matrix_L2}
\nonumber
U_{bfpm}(B) =
\begin{bmatrix}
0.0 & , & 0.5 & , & {\bf 1.0} & , & 0.5 & , & 0.0 \\
1.0 & , & 1.0 & , & {\bf 1.0} & , & 1.0 & , & 1.0\\
\end{bmatrix}
\end{equation}\\[0.2cm]
The origin can participate in all clusters (lines) as a full member. These objects such as the origin and alike are called \textit{critical} objects \cite{thirty two}. Addressing the critical objects is very important, as we need to encourage or prevent objects to/from participating in other clusters. Critical objects are important even in multi objective optimization problems, as we are interested to find the solution that fits in all objective functions. The arithmetic operations in set theory, the intersection $A \cap B$ and the union $A \cup B$, are precisely covered by BFPM. Covering the intersection by learning methods, removes the limitations and restrictions in membership assignments, and members can participate in more even all clusters.  \\ [0.3cm]
BFPM has the following properties:\\[0.2cm]
\textbf{Property 1}: \\[0.2cm]
\hspace*{5mm}Each data object must be assigned to at least one cluster.\\[0.4cm]
\textbf{Property 2}: \\[0.2cm]
\hspace*{5mm}Each data object can potentially obtain a membership value of 1 in multiple clusters, even in all clusters.\\[0.3cm]
\textbf{Proof of Property 1}: \\[0.2cm]
\hspace*{5mm}As the following inequality shows, each data object must participate in at least one cluster.
\begin{equation}
\nonumber
     0< \; 1/c \sum_{i=1}^c u_{ij}
\end{equation}
\textbf{Proof of Property 2}: \\[0.2cm]
\hspace*{5mm}According to fuzzy membership assignment \cite{thirty three} with respect to $c$ clusters, we have:
$  (0 < u_{1j} \leq 1, \;\;\; \forall \; x_j \in C_1 )$, ...,
$ (0 < u_{cj} \leq 1, \;\;\; \forall \; x_j \in C_c )$.
By replacing the values of zero and one (for fuzzy functions) with the linguistic values of $Z_m$ and $O_m$, respectively, we will have:
$(Z_m < u_{1j} \leq O_m, \;\;\;  \forall \; x_j \in C_1) $, ..., 
$ (Z_m < u_{cj} \leq O_m, \;\;\; \forall \; x_j \in C_c )$.\\
Consequently, we will get Eq. (\ref{total-set-mem-ling}), { as $Z_m$ and $O_m$ are upper and lower boundaries, and regarding the rules in fuzzy sets \cite{thirty four}, results from multiplications in upper and lower boundaries are in those boundaries}.\\
\begin{equation}
\nonumber
{ c.0 < \sum_{i=1}^cu_{ij} \leq c.1, \;\; \forall \; \; x_j \in C_i }\\
\end{equation}
\begin{equation}
\label{total-set-mem-ling}
c. Z_m < \sum_{i=1}^cu_{ij} \leq c.O_m, \;\; \forall \; \; x_j \in C_i \\
\end{equation}
By considering Eq. (\ref{total-set-mem-ling}), the above assumptions, and dividing all sides by $c$, we obtain the following equation: \\
\begin{equation}
\nonumber
{ 0 < \frac{1}{c} \sum_{i=1}^cu_{ij} \leq 1, \;\; \forall \; \; x_j \in C_i }\\
\end{equation}

\begin{equation}
\label{total-set-mem}
 Z_m < \frac{1}{c} \sum_{i=1}^cu_{ij} \leq O_m, \;\; \forall \; \; x_j \in C_i \\
\end{equation}

\begin{table*}[!ht]
\begin{center}
\caption{Analysing samples with regards to covariant variables five year survival(short/long), cancer stage(low/high), average age, and cancer type(s/a).}
\label{Cancer-T}
\begin{tabular}{ | c | c | c | c | c | c | c | c | c |    }
\hline
{\bf Serum metabolites} &  {\bf Samples No.} & {\bf
 Survival(s)} & {\bf Survival(l)} & {\bf Stage(l)} & {\bf Stage(h)} &  {\bf Ave. age} &  {\bf  C-type(s)} & {\bf C-type(a)}  \\
\hline
\hline 
 {\bf Diamond cluster } & 12 & 63.30 \% & 36.70 \% & 83.33 \%  & 16.67 \% & 64.82 &  57.33 \% & 42.67 \%  \\
\hline
{\bf Cancer samples in} & 21 & 40.91 \% & 59.09 \% & 60.87 \%  & 39.13 \% & 65.31 &  39.13 \% & 60.87 \%  \\
 {\bf square cluster} & &  &  &   &  &  &    &   \\
\hline
 {\bf Cancer samples in} & 6 & 50.00 \%  & 50.00 \% & 66.66 \%  & 33.34 \% & 58.60 &  33.34 \% & 66.66 \%  \\
 {\bf circle cluster} & &  &  &   &  &  &    &   \\
\hline
\end{tabular}
\end{center}
\end{table*}

\begin{figure*} [!ht]
\begin{center}
\leavevmode\fbox{\parbox[b][45mm][s]{140mm}{
\vfill\footnotesize {\includegraphics[width=14cm,height=4.5cm]{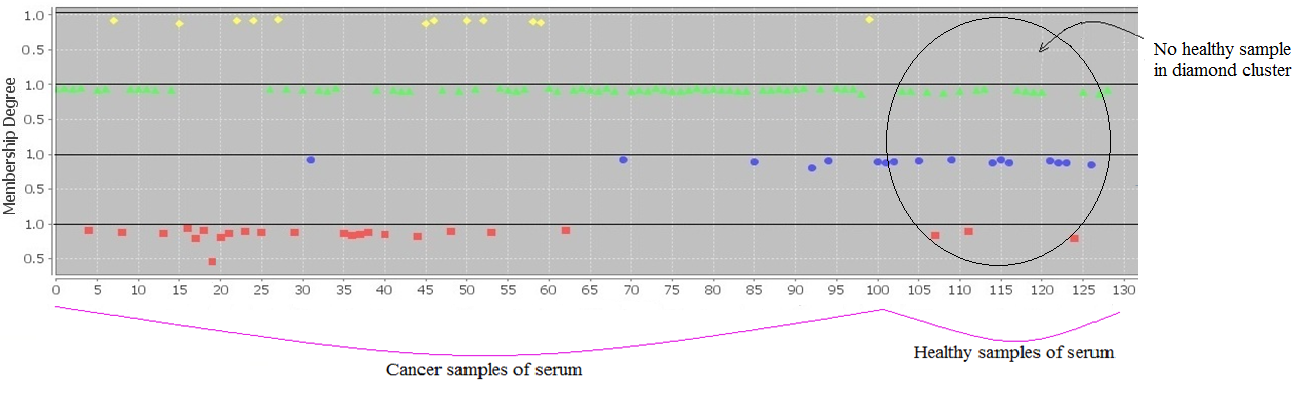}}\vfill}}
\caption{Serum samples categorized into 4 clusters to find a set of features that separate healthy samples from cancer samples, while there is no healthy sample in the diamond cluster.}
\label{Cancer-M}
\end{center}
\end{figure*}
Finally, we get the BFPM condition by converting the linguistic values to crisp values, $0 < 1/c \sum_{i=1}^c u_{ij} \leq 1, \;\; \forall j $. \\
The second proof provides the most flexible environment for all membership functions that make use of uncertainty in their membership assignments. In other words, BFPM is introduced to assist fuzzy and possibilistic methods in their membership assignments. In conclusion, BFPM presents a methodology that not only objects can be clustered based on their similarities, but also the abilities of objects in participation in other clusters can be studied. The method covers the intersection and the union operations with respect to all objects and clusters. It means that if and even if there is a similarity between any object and clusters, even in one dimension, the object can obtain membership degrees from clusters. If there is a similarity between an object and a cluster, a proper membership degree will be assigned to the object, otherwise the membership degree will be zero. The degree of membership is calculated based on the similarity of objects with clusters with respect to all dimensions. Having similarities in more dimensions results in a higher degree of membership. Moreover, the method considers the object's movement from one cluster to another in prediction and prevention strategies. In crucial systems such as security, diagnosing diseases, risk management, and decision making systems, studying the behaviour of objects in advance is extremely important. Neglecting the study of objects' movements directs the systems to irreparable consequences. Algorithm \ref{BFPCM} is introduced to assign memberships to objects with respect to each cluster.

\begin{algorithmic}
\begin{algorithm}
\caption{BFPM Algorithm}
\label{BFPCM}
\textbf{Input:\; X,} c, m	\\
\textbf{Output:\; U, V} 	
\STATE  \textbf{Initialize V;}
\WHILE {$ \underset{1 \leq  k  \leq c }{max} { \lbrace || V_{k,new} - V_{k,old} ||^2 \rbrace}  > \varepsilon$ }
    \STATE
    \begin{equation}
	\label{Fuzzy update}
		u_{ij} = \Big{[}\sum_{k=1}^c \big{(}\frac{||X_j - v_i||}{||X_j - v_k||} \big{)}^{\frac{2}{m-1}}\Big{]}^{\frac{1}{m}} , \;\; \forall i,j
	\end{equation}
	 \STATE
 	 \begin{equation}
	\label{prototype update}
	V_i = \frac{\sum_{j=1}^n (u_{ij})^m x_j}{ \sum_{j=1}^n (u_{ij})^m} ,\; \; \forall i \; ; \; \; \; \; \;  { (0 < \frac{1}{c} \sum_{i=1}^c u_{ij} \leq 1)}.
	\end{equation}
\ENDWHILE
\end{algorithm}
\end{algorithmic}

 BFPM methodology and Euclidean distance function have been applied in Algorithm \ref{BFPCM}. This algorithm makes use of the BFPM membership assignments (Eq. (\ref{flexible fuzzy partition})) by considering ${(0 < \frac{1}{c} \sum_{i=1}^c u_{ij} \leq 1)}$ condition to assign membership values. Eq. (\ref{Fuzzy update}) and Eq. (\ref{prototype update}) show how the algorithm calculates $(u_{ij})$ and how the prototypes $(v_i)$ will be updated in each iteration. The algorithm runs until reaching the condition:\\
\begin{equation}
\nonumber
 \max_{1 \leq  k  \leq c } { \lbrace || V_{k,new} - V_{k,old} ||^2 \rbrace}  < \varepsilon
 \end{equation}
 The value assigned to $\varepsilon $ is a predetermined constant that varies based on the type of objects and clustering problems.
%
%
\begin{table*}[!ht]
\caption{Accuracy rates of modified methods: k-means, FCM, and PCM in comparison with BFPM for Iris dataset. }
\label{Accuracy-Iris-Unsupervised}
\begin{center}
\begin{tabular}{  c  c  c  c  c  c  c  c  c  c  c}
\hline
\multicolumn{10}{c}{{\color{white} \rot{.........}} \bf Iris Dataset}\\[0.1ex]
\hline
{\color{white} \rot{.........}}& \multicolumn{7}{c}{\small \bf Modified FCM} &  & {\small \bf PCM} & \\[0.1ex] 
\cline{2-8}
\cline{10-10}
 {\small \rot{Fuzzy} \rot{k-means}}
 & {\small \rot{LAC}} & { \small \rot{WLAC}} & {\small \rot{FWLAC}} & {\small \rot{CD-FCM}}  & {\small \rot{KCD-FCM }} &  {\small \rot{KFCM-F}} & {\small \rot{WE-FCM}} &  & {\small \rot{APCM}} &  {\small \rot{BFPM}} \\ [0.1ex]
\hline
{ {\color{white}\rot{.........}}}{ 74.96} & { 90.21}  & { 90.57} & { 94.37} & { 95.90}  & { 96.18}  & { 92.06} & { 96.66} &  & { 92.67} & { { 97.33}}  \\[0.1ex]
\hline
\end{tabular}
\end{center}
\end{table*}
%
\begin{figure*} [!ht]
\begin{center}
\leavevmode\fbox{\parbox[b][45mm][s]{140mm}{
\vfill\footnotesize {\includegraphics[width=14cm,height=4.5cm]{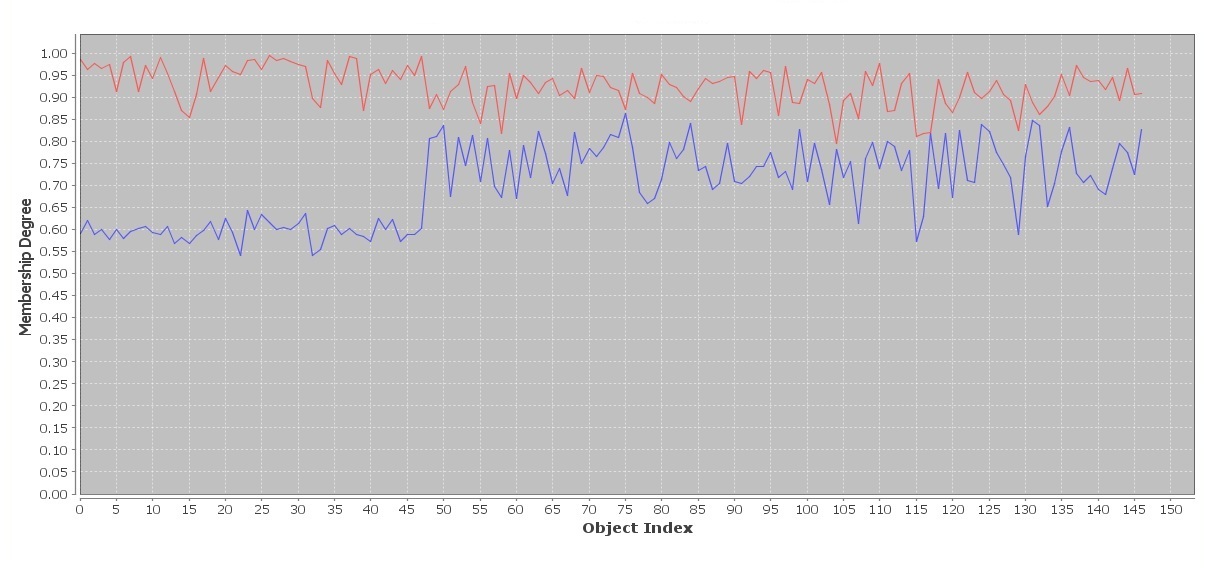}}\vfill}}
\caption{Plot obtained by BFPM to demonstrate objects' memberships with respect to two clusters, the current and the closest clusters, to analyse objects' movements from their own cluster to another. }
\label{Iris-M}
\end{center}
\end{figure*}
\begin{figure*} [!ht]
\begin{center}
\leavevmode\fbox{\parbox[b][45mm][s]{140mm}{
\vfill\footnotesize {\includegraphics[width=14cm,height=4.5cm]{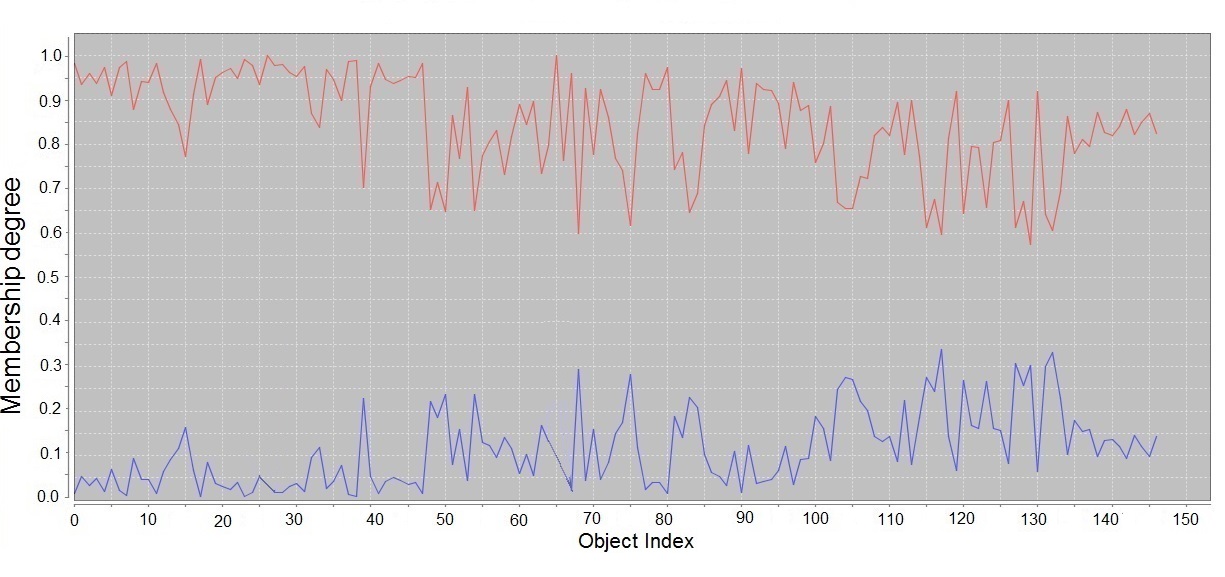}}\vfill}}
\caption{Plot obtained by fuzzy methods to demonstrate objects' memberships with respect to two clusters, the current and the closest clusters.}
\label{Iris-F}
\end{center}
\end{figure*}
\section{\textbf{Experimental Verification}}
\label{Exp_result}
The proposed method has been applied on real applications and problems from different domains such as medicine (lung cancer diagnosis) \cite{thirty}, intrusion detection systems \cite{thirty two}, and banking (financial) systems and risk managements (decision making systems) \cite{thirty five}. The proposed methodology has been utilized in both supervised and unsupervised learning strategies, but in this paper clustering problems have been selected for experimental verifications. BFPM revealed information about lung cancer through analysis of metabolomics, by evaluating the potential abilities of objects to participate in other cluster. The results on the cancer dataset have been presented by Table \ref{Cancer-T} and Fig. \ref{Cancer-M}. The figure shows how individuals are clustered with respect to their serum features (metabolites), where the horizontal axis presents objects, and the vertical axis shows the memberships assigned to objects. The methodology has been also applied in risk management and security systems to evaluate how objects (packets/transactions) that are categorized in normal cluster can be risky for the systems in the near future. In other words, the potential ability of current normal (healthy) objects to move to abnormal (cancer) cluster, and vice versa, has been covered by BFPM. In this paper, the benchmark Iris and Pima datasets from UCI repository \cite{thirty six} have been chosen to illustrate the idea. The datasets used in this experiment were normalized in the range $[0,1]$. The accuracy of BFPM is compared with other methods, while the accuracy is mostly measured by the percentage of the correct labelled objects in classification problems, however the accuracy in clustering problems refers to the evaluation of the distance between the objects and the center of the clusters which is known as measuring separation and compactness of objects with respect to prototypes \cite{thirty seven}. Table \ref{Accuracy-Iris-Unsupervised} shows the accuracy obtained by BFPM in comparison with recent fuzzy and possibilistic methods: fuzzy-k-means \cite{thirty eight}, Locally Adaptive Clustering (LAC) \cite{thirty eight}, Weighted Locally Adaptive Clustering (WLAC) \cite{thirty eight}, Fuzzy Weighted Locally Adaptive Clustering (FWLAC) \cite{thirty eight}, Collaborative Distributed Fuzzy C-Means (CD-FCM) \cite{thirty nine}, Kernel-based Collaborative Distributed Fuzzy C-Means (KCD-FCM) \cite{thirty nine}, Kernel-based Fuzzy C-Means and Fuzzy clustering (KFCM-F) \cite{thirty nine}, Weighted Entropy-regularized Fuzzy C-Means (WE-FCM) \cite{thirty nine}, and Adaptive Possibilistic C-Means (APCM) \cite{forty}. Results form BFPM was achieved by assigning fuzzification constant as ($m=2$).
%
 \begin{table*}[!t]
\caption{Validity Indices : Partition Coefficient($ V_{pc}$), Partition Entropy ($ V_{pe}$), DB, CS, and  G index with their functions and desirable values, for $n$ data objects with respect to $c$ clusters.}
\label{Validity Functions}
\begin{center}
\begin{tabular}{ | c | c | c |  }
\hline
Validity Index & Suitable Value & Function  \\
\hline
\hline
$V_{pc} $ & Maximum  &  \parbox{12cm}{ \begin{equation} \label{VPC(u)}  V_{pc}(U) = \frac{\sum_{j=1}^n \sum_{i=1}^c(u_{ij}^2)}{n} \end{equation}}\\
& & $u_{ij}$ is the membership value of $j^{th}$ object for $i^{th}$ cluster. \\
\hline
 $V_{pe}$ & Minimum  & \parbox{12cm}{\begin{equation} \label{VPE(u)}  V_{pe}(U) =\frac{-1}{n} \bigg\lbrace \sum_{j=1}^n \sum_{i=1}^c(u_{ij} \;  log \;u_{ij})\bigg\rbrace \end{equation}}\\
 \hline
 DB & Minimum  & \parbox{12cm}{\begin{equation} \label{DB} DB(C) = \frac{1}{C} \sum_{j=1}^C R_j  \; ; \; \;\; R_j = max_{j,j\neq i} \left( \frac{ e_i + e_j}{D_{ij}} \right)  \;  ; \;\;\; e_j = \frac{1}{N_j} \sum_{x\in C_j} || X-m_j||^2 \end{equation}} \\
   &  & $D_{ij}$ is the distance between the $i^{th}$ and the $j^{th}$ prototype \\
   &  & $e_i$ and $e_j$ are the average errors for clusters $C_i$ and $C_j$. \\
\hline
CS &  Minimum  &  \parbox{12cm}{\begin{equation} \label{CS} CS(C) = \frac{\sum_{j=1}^C \bigg(\frac{1}{N_j} \sum_{X_i \in C_j} (max_{x_l \in C_j} D(X_l,X_j))\bigg)}{\sum_{j=1}^K \bigg( max_{j\in K,j \neq i} D(m_i,m_j) \bigg)} \end{equation}}\\
\hline 
  G & Maximum  & \parbox{12cm}{\begin{equation}  
\label{G}
G = \frac{D}{C} =  \frac{\frac{1}{n^2} \sum_{j_1=1}^n \sum_{j_2=1}^n d^2 \big(X_{j_1},X_{j_2} \big)w_2}{\frac{2}{n(n-1)} \sum_{j_1=1}^{n-1} \sum_{j_2=j_1+1}^n \sum_{i=1}^c d^2 \big(X_{j_1},X_{j_2}\big) w_1}
\end{equation} }\\
&  &  $w_2 = min \lbrace max_{j_1}u_{i_1 j_1} , max_{j_2 \neq j_1} u_{i_2 j_2} \rbrace, $ and  $  w_1 = min \lbrace u_{i_1j} , u_{i_2j} \rbrace$\\ 
\hline
\end{tabular}
\end{center}
\end{table*}

\begin{table*}[!ht]
\caption{Results of BFPM algorithm, using $V_{pc}$, $V_{pe}$, $DB$, $G$, and $CS$ indices as fitness functions for normalized Iris and Pima datasets for different values of fuzzification constant $(m)$. }
\label{Val_clus_No}
\begin{center}
\begin{tabular}{ | c | c | c | p{1cm} | p{1cm} | p{1cm} | p{1cm} |   }
\hline
Validity Index & DataSet & Cluster No. &  m=1.4 & m=1.6 & m=1.8 & m =2 \\
\hline
\hline
\multirow{2}{*}{\textbf{$V_{pc_{\uparrow}}$}}		
		&Iris	& 3 &   0.761 & 0.744 & 0.739 &  0.742\\	
	\cline{3-7}
		&  Pima		& 2 &   0.758 & 0.660 & 0.574& 0.573\\	
		
		\hline
		\hline
\multirow{2}{*}{\textbf{$V_{pe_{  \downarrow}}$}}		
		& Iris 	& 3 &   0.332 & 0.271 & 0.233 & 0.204 \\	
		\cline{3-7}
		& Pima 		& 2 &  0.301 & 0.301 & 0.296& 0.268\\	
\hline
\hline
\multirow{2}{*}{\textbf{$DB_{  \downarrow}$}}
		&  Iris	& 3    & 0.300  & 0.249  & 0.232 & 0.225  \\	
	\cline{3-7}
		& 	Pima	& 2  & 2.970 & 1.950  & 1.742  & 1.668  \\
		
\hline
\hline
\multirow{2}{*}{\textbf{$G_{\uparrow}$}}
	&  Iris	& 3  & 5.300 & 7.880  & 10.050 & 12.140  \\	
		\cline{3-7}
		& 	Pima	& 2  & 1.440 & 1.830 & 2.100 & 2.230   \\
\hline
\hline		
\multirow{2}{*}{\textbf{$CS_{  \downarrow}$ }}
		&  	Iris	& 3  & 0.051 & 0.047  & 0.046  & 0.045 \\	
		\cline{3-7}
		& Pima		& 2  & 0.054 & 0.036 & 0.032 & 0.031    \\
		
\hline
\end{tabular}
\end{center}
\end{table*}
According to the results, BFPM performs better than other clustering methods, in addition to provide the crucial objects and areas in each dataset by allowing objects to show their potential abilities to participate in other clusters. This ability results in tracking the objects' movements from one cluster to another.
Fig. \ref{Iris-M} and Fig \ref{Iris-F} depict the objects' memberships obtained by BFPM and fuzzy methods, respectively. The memberships are assigned with respect to two clusters, the current cluster which objects are clustered in and the closest clusters. The horizontal axis presents objects, and the vertical axis shows the memberships assigned to objects. The upper points are memberships assigned to objects with respect to the current cluster and the lower points depicts the objects' memberships with respect to the closest cluster. According to Fig. \ref{Iris-M}, the objects can show their ability to participate in other clusters by obtaining higher memberships based on BFPM membership assignments $(0 < \frac{1}{c} \sum_{i=1}^c u_{ij} \leq 1)$, while in Fig. \ref{Iris-F} which is obtained by fuzzy methods, objects cannot obtain high memberships for other clusters as the fuzzy methods are designed to get objects completely separated $(\sum_{i=1}^c u_{ij} =1)$. By comparing the figures, we can conclude that fuzzy methods aim to cluster data objects, while BFPM not only aims to cluster objects with higher accuracy but also detect critical objects that trap learning methods in their learning procedures. The accuracy of clustering methods can be also evaluated either using cluster validity functions or the comparison indices method \cite{forty one}. Several validity functions have been introduced, which some of them are presented in Table \ref{Validity Functions}. DB index \cite{forty two}, shown by Eq. (\ref{DB}), evaluates the performance of the clustering method by maximizing the distance between prototypes distances on one side and minimizing the distance between each prototype and the objects belong to the same cluster. CS index, presented by Eq. (\ref{CS}) \cite{forty three}, is very similar to DB index, but it works on clusters with different densities. G index, presented by Eq. (\ref{G}) \cite{forty four}, performs by evaluating the separations and compactness of data objects with respect to clusters. Separation of the fuzzy partition is defined as $D$ to check how well the prototypes are separated. On the other hand, compactness $C$ of fuzzy partitions measures how close data objects are in each  cluster. The desirable values ($ min / max $) with respect to each validity index is presented in Table \ref{Validity Functions}. Table \ref{Val_clus_No} explores values from different validity functions $V_{pc}$, $V_{pe}$, $DB$, $G$, and $CS$ for BFPM with respect to different values of fuzzification constant $m$ for Iris and Pima datasets with three and two clusters respectively.
\section{\textbf{Discussions and Conclusions}}
\label{Conclusion}
This paper introduced the Bounded Fuzzy Possibilistic Method (BFPM) as a new methodology in membership assignments in partitioning methods. The paper provided the mathematical proofs for presenting BFPM as a superset of conventional methods in membership assignments. BFPM not only avoids decreasing the memberships assigned to objects with respect to all clusters, but also makes the search space wider for objects to participate in more, even all clusters as partial or full members to present their potential abilities to move from one cluster to another (mutation). BFPM facilitates the analysis of objects' movements for crucial systems, while conventional methods aim to just cluster objects without paying attention to their movements. Tracking the behavior of objects  in advance leads to better performances, in addition to have a better insight in prevention and prediction strategies. The necessity of considering the proposed method has been proved by several examples from geometry, set theory, and other domains.

\begin{IEEEbiography}[{\includegraphics[width=1in,height=1.25in,clip,keepaspectratio]{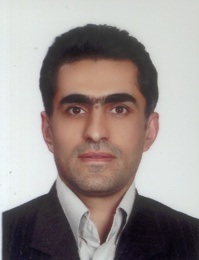}}]{Hossein Yazdani} is a PhD candidate at Wroclaw University of Science and Technology. He was a Manager of Foreign Affairs, DBA, Network Manager, System Analyst in Sadad Informatics Corp. a subsidiary of Melli Bank of Iran. Currently, he cooperates with the Department of Information Systems, Faculty of Computer Science and Management, and Faculty of Electronics at Wroclaw University of Science and Technology. His research interests include BFPM, Critical objects, Machine learning, Artificial intelligence, Dominant features, Distributed networks, Collaborative clustering, Security, Big data, Bioinformatics, and optimization.
\end{IEEEbiography}
\end{document}